\documentclass[sigconf]{acmart}
\AtBeginDocument{%
  }

\setcopyright{acmlicensed}
\copyrightyear{2026}
\acmYear{2026}
\setcopyright{cc}
\setcctype{by}
\acmConference[SIGIR '26]{Proceedings of the 49th International ACM SIGIR Conference on Research and Development in Information Retrieval}{July 20--24, 2026}{Melbourne, VIC, Australia}
\acmBooktitle{Proceedings of the 49th International ACM SIGIR Conference on Research and Development in Information Retrieval (SIGIR '26), July 20--24, 2026, Melbourne, VIC, Australia}
\acmDOI{10.1145/3805712.3809737}
\acmISBN{979-8-4007-2599-9/2026/07}

\usepackage{amsmath} 
\usepackage{amsthm}
\usepackage{enumitem}
\usepackage{multirow}
\usepackage{booktabs}
\usepackage{makecell}  
\usepackage{array} 
\usepackage{tabularx}
\usepackage{graphicx}
\usepackage[table]{xcolor}
\usepackage{tcolorbox}
\tcbuselibrary{listings}
\tcbuselibrary{breakable}
\definecolor{purple}{HTML}{D8ABCF}  
\definecolor{pink}{HTML}{F0D0E7}  
\begin{document}

\title{Deja Vu in Plots: Leveraging Cross-Session Evidence with Retrieval-Augmented LLMs for Live Streaming Risk Assessment}


\author{Yiran Qiao}
\authornote{This work was conducted during Yiran's internship at ByteDance China.}
\authornotemark[3]
\authornotemark[4]
\orcid{0009-0006-0632-0066}
\affiliation{%
  \institution{Institute of Computing Technology, Chinese Academy of Sciences}
  \city{Beijing}
  \country{China}
}
\email{yrqiao@gmail.com}

\author{Xiang Ao}
\authornote{Corresponding Author.}
\authornotemark[3]
\authornotemark[4]
\orcid{0000-0001-9633-8361}
\affiliation{%
  \institution{Institute of Computing Technology, Chinese Academy of Sciences}
  \city{Beijing}
  \country{China}
}
\email{aoxiang@ict.ac.cn}

\author{Jing Chen}
\orcid{0000-0003-2672-4587}
\affiliation{%
  \institution{ByteDance China}
  \city{Hangzhou}
  \country{China}
}
\email{yilan.chan@bytedance.com}

 \author{Yang Liu}
 \authornotemark[3]
\authornotemark[4]
 \orcid{0000-0002-1525-0788}
\affiliation{%
  \institution{Institute of Computing Technology, Chinese Academy of Sciences}
  \city{Beijing}
  \country{China}
}
\email{liuyang2023@ict.ac.cn}

\author{Qiwei Zhong}
\orcid{0000-0002-8517-8072}
\affiliation{%
  \institution{ByteDance China}
  \city{Hangzhou}
  \country{China}
}
\email{huafeng.hf@bytedance.com}

 \author{Qing He}
\authornote{State Key Laboratory of AI Safety, Institute of Computing Technology, Chinese Academy of Sciences.}
\authornote{Also with University of Chinese Academy of Sciences.}
 \orcid{0000-0001-8833-5398}
\affiliation{%
  \institution{Institute of Computing Technology, Chinese Academy of Sciences}
  \city{Beijing}
  \country{China}
}
\email{heqing@ict.ac.cn}
\renewcommand{\shortauthors}{Yiran Qiao et al.}

\begin{abstract}
The rise of live streaming has transformed online interaction, enabling massive real-time engagement but also exposing platforms to complex risks such as scams and coordinated malicious behaviors. Detecting these risks is challenging because harmful actions often accumulate gradually and recur across seemingly unrelated streams. To address this, we propose \textbf{CS-VAR} (Cross-Session Evidence-Aware Retrieval-Augmented Detector) for live streaming risk assessment. In CS-VAR, a lightweight, domain-specific model performs fast session-level risk inference, guided during training by a Large Language Model (LLM) that reasons over retrieved cross-session behavioral evidence and transfers its local-to-global insights to the small model. This design enables the small model to recognize recurring patterns across streams, perform structured risk assessment, and maintain efficiency for real-time deployment. Extensive offline experiments on large-scale industrial datasets, combined with online validation, demonstrate the state-of-the-art performance of CS-VAR. 
Furthermore, CS-VAR provides interpretable, localized signals that effectively empower real-world moderation for live streaming.
\end{abstract}




\maketitle

\section{Introduction}
Live streaming has exploded into a global phenomenon on the Web, transforming how we connect, communicate, and consume content. This vibrant ecosystem combines entertainment, social interaction, and commerce, creating unprecedented opportunities for engagement~\cite{hilvert2018social,chen2024would,wang2022live,qiao2026deja}.
%with the global market share expected to reach \$345.13 billion by 2030. 
%Managing such a complex application requires understanding the behavioral traces left by millions of users.

Each live session features dense, high-frequency interactions between hosts and viewers, from chat messages to virtual gifts. While these interactions drive engagement, they can also be exploited for coordinated scams. Even focusing only on a single platform's e-commerce live streaming, the scale of misconduct is striking: over 50,000 hosts were sanctioned between 01/2025 and 08/2025\footnote{\url{https://finance.sina.com.cn/tech/2025-08-22/doc-infmvzec3584172.shtml}}. This prevalence underscores the critical need for robust risk assessment.

The challenge of risk assessment in this domain is twofold and highly interconnected. First, risks are inherently \textbf{coordinated and real-time}, building gradually over time. A single harmful action, though seemingly minor, can contribute to a progressively accumulating risk pattern as it combines with other interdependent user behaviors. This dynamic buildup demands a fine-grained approach that can capture the full spatiotemporal accumulation of user interactions. To facilitate such fine-grained modeling, it is useful to conceptualize continuous user activities as short, semantically coherent behavioral segments, which can later serve as basic units for detailed analysis.

\begin{figure}[t]
\centering
\includegraphics[width=0.5\textwidth]{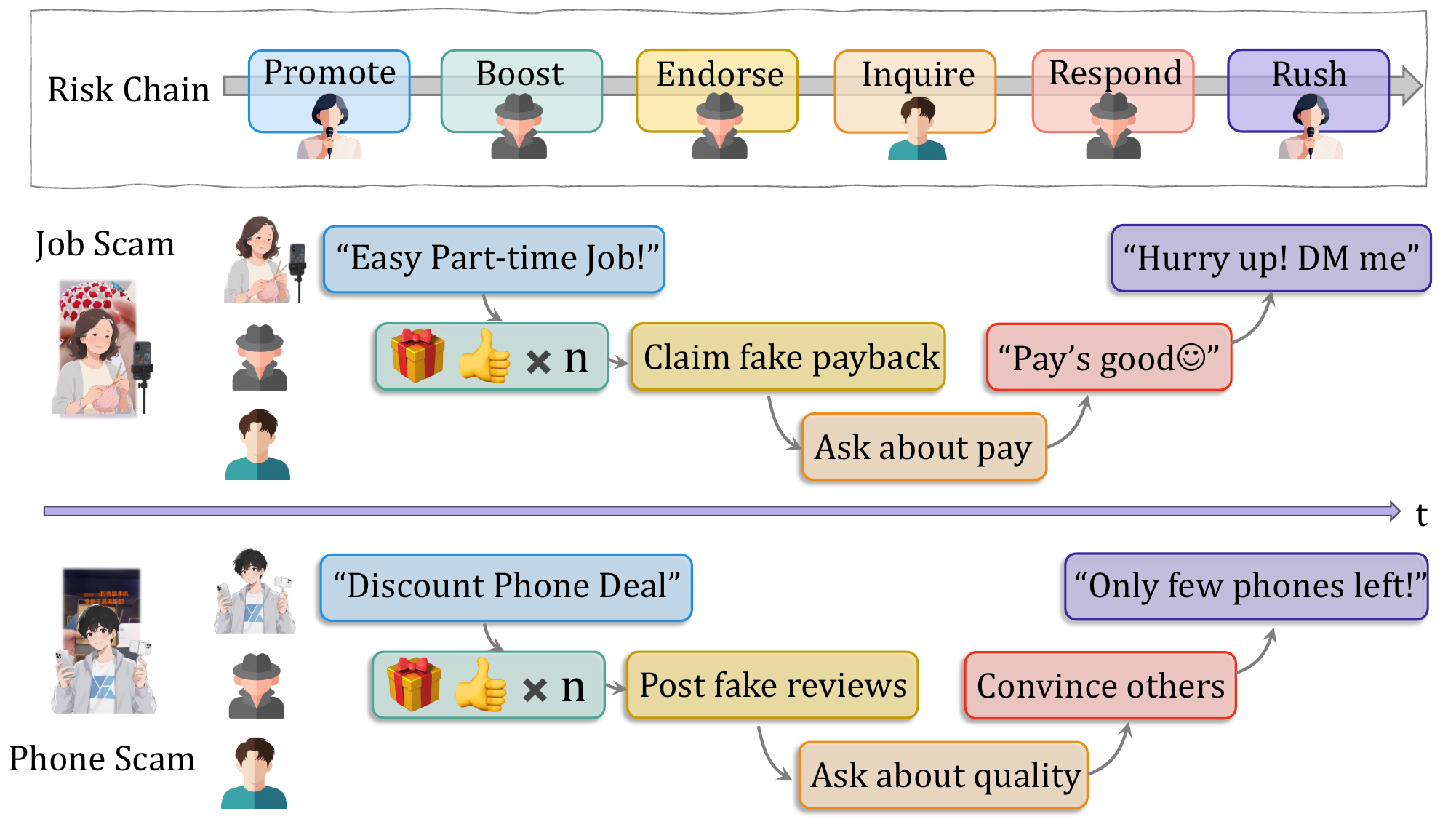}
\caption{A toy example illustrating behavioral patch chains in two distinct scam scenarios (part-time job scam vs. cheap phone scam) involving hosts, representative shills, and victims-to-be. Despite the different surface contexts, both follow a nearly identical progression of patches, from promotion to rushing audience.}
\label{fig:intro}
\end{figure}

This granular perspective on user behavior reveals our second key insight: malicious activities often form \textbf{similar risk chains across different streams}. In practice, we observe cases where streams that appear unrelated, such as those advertising part-time jobs or selling discounted phones, nonetheless exhibit nearly identical scam routines~(c.f.~Figure~\ref{fig:intro}). These chains of behavioral segments naturally motivate reasoning at a patch-level scale: by treating short, coherent segments as retrieval units, one can aggregate cross-session evidence to surface recurring malicious patterns.

However, the effectiveness of retrieval at the patch level hinges on its ability to learn robust representations of malicious patterns, which requires a massive amount of fine-grained ground truth data. Since obtaining such detailed annotation is infeasible in practice, we introduce a retrieval-augmented framework that bypasses this supervision bottleneck. Our approach leverages the powerful reasoning capabilities of Large Language Models~(LLMs)~\cite{brown2020language,zhao2023survey,pan2024unifying}, using retrieved cross-session evidence as a form of contextual supervision. This enables effective risk assessment by allowing the LLM to cross-check and reason about a stream’s overall risk without relying on extensive, human-annotated labels.

Despite their strong reasoning capabilities, directly applying LLMs to live risk detection is impractical. Their massive scale incurs prohibitive~\cite{miao2025towards}, and their limited context window~\cite{patil2024review} is easily saturated by the long and noisy behavioral streams in live sessions. This gap makes it infeasible to rely on LLMs alone for timely and scalable risk control in practice.
What is needed is a paradigm that reconciles two conflicting demands: the ability to reason over distributed evidence and the efficiency required for real-time deployment. We achieve this by coupling a domain-specific small model with an LLM, where the former ensures fast inference while the latter provides rich cross-session reasoning during training.

Building on this idea, we propose a \textbf{C}ross-\textbf{S}ession E\textbf{V}idence-\textbf{A}ware \textbf{R}etrieval-Augmented Detector~(abbr. \textbf{CS-VAR}) for live streaming risk assessment. In CS-VAR, a domain-specific small model acts as a front-end, providing the LLM reasoning with both key patches and the basis embeddings for cross-session retrieval. In turn, the LLM's retrieval-augmented reasoning is distilled back into the small model, empowering it to perform fast, independent inference and meet the demanding timeliness of real-time risk detection. Through this design, CS-VAR unifies local behavioral cues with cross-session patterns into coherent reasoning chains, combining fine-grained interpretability with scalability for industrial deployment. %Note that we use LLMs not as generators, but as a scalable surrogate annotator for cross-session relevance signals.

Extensive offline and online evaluations confirm that CS-VAR not only achieves state-of-the-art performance but also generates patch-level signals that support actionable moderation. 
%Since November 2025, it has been deployed on a major live streaming platform for large-scale risk assessment.
Our main contributions are summarized as follows:
\begin{itemize}[leftmargin=*,topsep=5pt]
\item We present, to the best of our knowledge, the first framework that leverages LLMs for session-level risk assessment in live streaming, where recurring cross-session routines make structured reasoning particularly effective.
\item We propose CS-VAR, which retrieves cross-session evidence to support LLM reasoning and distills the resulting multi-granularity judgments into a lightweight model, achieving both cross-session awareness and real-time efficiency.
\item Extensive offline experiments and large-scale deployment on a major live streaming platform confirm that CS-VAR delivers state-of-the-art performance and provides interpretable, patch-level cues that enhance actionable content moderation.
\end{itemize}
\section{Related Work}
%\noindent \textbf{User Behavior Modeling for Risk Control.}
\subsection{User Behavior Modeling for Risk Control}
Modeling user behavior is central to risk assessment, particularly for detecting subtle or coordinated malicious activities. Traditional approaches often rely on rule-based or statistical models~\cite{dimitras1996survey,chen2016financial}, which are limited in capturing complex temporal and relational patterns. More recent methods adopt sequential~\cite{branco2020interleaved,liu2020fraud,qiao2024Financial} or graph-based~\cite{huang2022auc,liu2021pick,shi2022h2,cheng2025graph,li2021live} representations to model user interactions over time. Retrieval- or memory-based mechanisms have also been explored to enrich representations with historical interactions or similar sessions and mitigate distribution shifts~\cite{qiao2025online,xiao2024vecaug}. For example, LIFE~\cite{li2021live} detects fraudulent transactions in e-commerce live streaming, still following a typical transaction-level detection setting, using only behavioral data from the stream.

In contrast, live streaming risk often emerges at the session level, reflecting aggregated patterns of host–viewer interactions~\cite{qiao2026live}. Sessions are dense and high-frequency, and malicious behaviors may recur across streams, necessitating prediction approaches that capture both within-session dynamics and cross-session patterns.
Such traits motivate retrieval-augmented reasoning with LLMs, allowing cross-session evidence to inform more robust risk assessment.

%\noindent \textbf{Retrieval-Augmented LLM Reasoning.} 
\subsection{Retrieval-Augmented LLM Reasoning}
Retrieval-Augmented Generation (RAG) was initially proposed to improve knowledge-intensive tasks by retrieving relevant documents from an external corpus before conditioning the LLM’s output~\cite{lewis2020retrieval,fan2024survey,gao2023retrieval}. By grounding predictions in retrieved evidence, RAG enhances accuracy, reduces hallucination, and increases interpretability compared to purely parametric models. Early work focused on generative tasks such as open-domain question answering \cite{izacard2021leveraging,fan2024survey}, while recent studies have extended RAG to discriminative tasks involving reasoning over sparse but informative signals \cite{shi2024replug, qiang2025exploring}. These studies demonstrate how retrieval can help LLMs focus on relevant evidence and produce more reliable outputs.

Unlike generic RAG pipelines that index raw text or require ad hoc chunking, our framework leverages small-model-derived embeddings that naturally decompose sessions into semantically coherent behavioral patches. Combined with session-aware summaries, this design grounds retrieval in evidence aligned with the small model’s risk-sensitive inductive bias, supporting interpretable cross-session reasoning.

\section{Problem Formulation}

\subsection{Business Setting}

Live streaming platforms face significant risks from fraudulent activities, illicit promotions, and manipulative behaviors that can cause financial and reputational damage. Effective risk detection must operate under several practical constraints:

\begin{itemize}[leftmargin=*,topsep=3pt]
    \item \textbf{Weak supervision}: Only session-level risk labels are available due to the high cost of fine-grained annotation.
    
    \item \textbf{Interpretability}: Operators require explainable evidence to support moderation decisions and regulatory compliance.
    
    \item \textbf{Scalability}: Systems must efficiently process thousands of concurrent sessions in real-time.
    
    \item \textbf{Recurring patterns}: Risky behaviors often follow similar scripted schemes across multiple sessions, allowing patterns to be leveraged for early detection.
\end{itemize}
These constraints motivate representing sessions as user-timeslot patch grids, which capture local behavioral patterns while supporting session-level risk assessment.

\subsection{Definition and Objective}
We study \emph{live streaming risk assessment}, which aims to determine whether a live streaming session involves risky behaviors such as fraud or illicit promotion. 
To formalize the problem, we adopt a grid-based abstraction of user actions (see Appendix~\ref{sec:app:notation} for a complete list of symbols and definitions).

\begin{definition}
\textbf{(Action)}  
An action in a live streaming session is represented as a tuple
\[
\alpha = (u, t, a, x),
\]
where $u$ is the user performing the action, $t$ is the timestamp, $a$ denotes the action type, and $x \in \mathbb{R}^d$ denotes the text embedding extracted from the raw text by a pretrained LM, while the raw form is retained for prompt construction.

\end{definition}

\begin{definition}
\textbf{(Live Streaming Session)}  
A session $s$ during time window $[0,T]$ is defined as
\[
S_s^{[0,T]} = \big(\mathcal{U}_s,  [\alpha_1, \alpha_2, \ldots, \alpha_{N_s}]\big),
\]
where the user set $\mathcal{U}_s = \{ u_s^{\mathrm{h}} \} \cup U_s^{\mathrm{v},[0,T]}$ consists of the unique host $u_s^{\mathrm{h}}$ and the participating viewers, and $[\alpha_1, \alpha_2, \ldots, \alpha_{N_s}]$ is the chronologically ordered sequence of $N_s$ actions within $[0,T]$.
Throughout the following, we drop the subscript $s$ when focusing on a single session for clarity.
\end{definition}

\begin{definition}
\textbf{(Temporal Discretization)}  
We partition $[0,T]$ into $K$ consecutive, non-overlapping timeslots $\{ \mathcal{T}_k \}_{k=1}^K$, where $\mathcal{T}_k = [(k-1)\Delta t,\, k\Delta t)$ and $\Delta t = T/K$.
\end{definition}

\begin{definition}
\textbf{(Patch)}  
For a user $u \in \mathcal{U}_s$ and timeslot $\mathcal{T}_k$, the corresponding patch is defined as
\[
p_{u,k} = [\, \alpha_{i_1}, \alpha_{i_2}, \ldots, \alpha_{i_{n_{u,k}}} \,],
\]
where $\alpha_{i_j} = (u, t_{i_j}, a_{i_j}, x_{i_j})$ with $t_{i_j} \in \mathcal{T}_k$ and $t_{i_1} < \cdots < t_{i_{n_{u,k}}}$. Here $n_{u,k}$ is the number of actions performed by $u$ in slot $k$. 
\end{definition}

\begin{definition}
\textbf{(Patch Grid)}  
The collection of all user-timeslot patches constitutes the patch grid of session $s$:
\[
\mathcal{P}_s = \{\, p_{u,k} \mid u \in \mathcal{U}_s, \; k = 1,\ldots,K \,\}.
\]
This grid provides a structured view of session dynamics, jointly capturing the spatial dimension (users) and temporal dimension (timeslots), and serves as the basic modeling unit for risk analysis.
\end{definition}

\noindent
\textbf{Problem Objective.}  
Given a dataset $\mathcal{D} = \{ (S_s^{[0,T]}, y_s) \}_{s=1}^N$ where $y_s \in \{0,1\}$ indicates whether session $s$ is risky~($y_s=1$), the task is to learn a function
\[
f: S_s^{[0,T]} \mapsto [0,1],
\]
that estimates the probability of risk for each session, supporting session-level predictions with patch-level signals provided.

\section{Methodology}

\begin{figure*}[tbp]
  \centering
  \includegraphics[width=\textwidth]{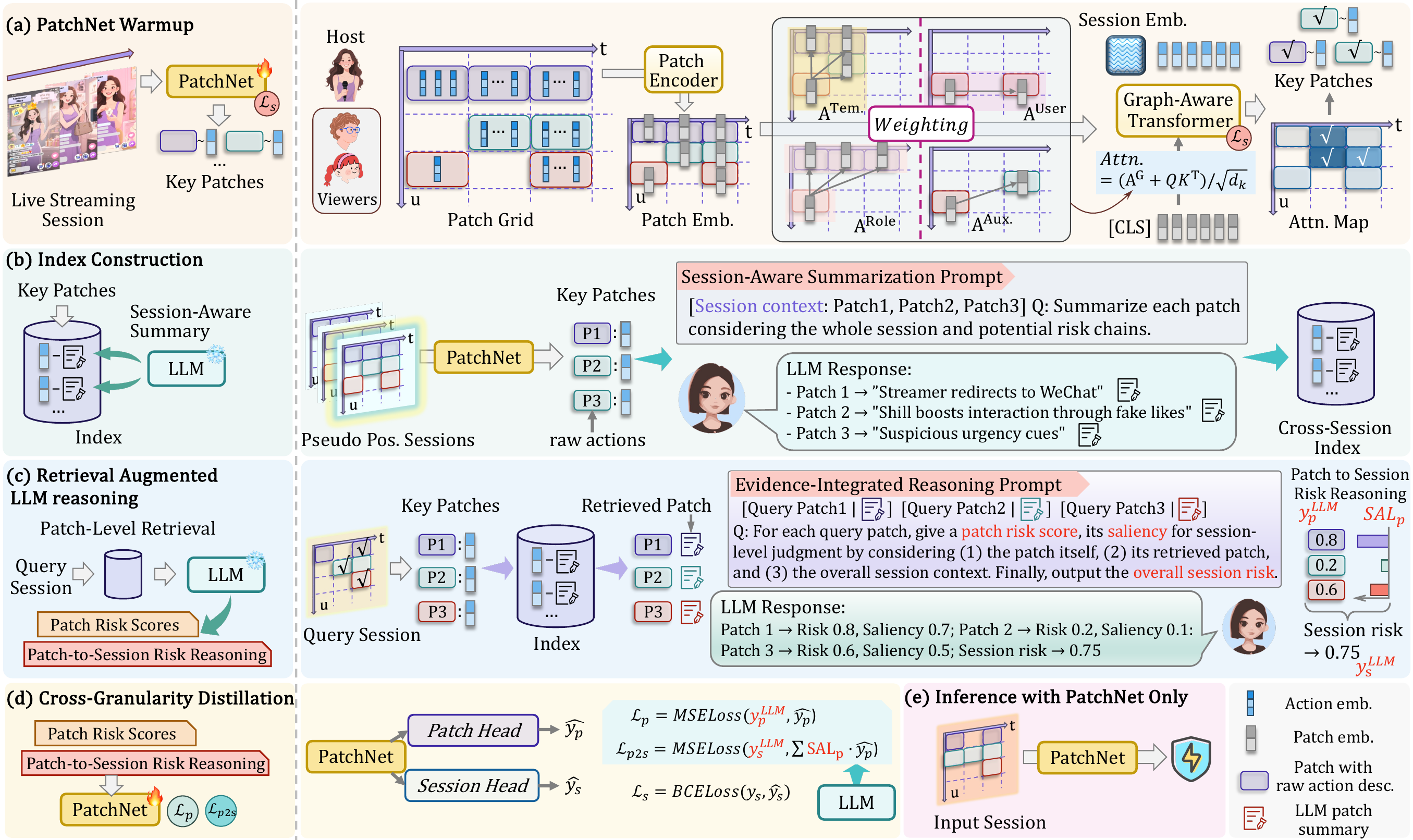}

  \caption{CS-VAR addresses live streaming risk detection by coupling lightweight PatchNet with retrieval-augmented LLM reasoning. (a) PatchNet highlights high-attention patches and (b) builds an index enriched with LLM-generated session-aware summaries. (c) Cross-session retrieval provides external reference signals, enabling the LLM to infer risks from patch to session level. (d) Distillation injects this reasoning chain into PatchNet, aligning patch evidence with overall session risk. (e) At deployment, PatchNet alone supports real-time, interpretable monitoring of live streams.
 }
  \label{fig:overview}
\end{figure*}
\subsection{Overview of CS-VAR}
%Figure~\ref{fig:overview} presents the overall framework of CS-VAR (Cross-Session Evidence-Aware Retrieval-Augmented Detector) for live streaming risk assessment. The design follows a coarse-to-fine paradigm where a domain-tailored lightweight model~(PatchNet) and a retrieval-augmented LLM collaborate in complementary roles. In \textbf{(a) PatchNet Warm-up}, the lightweight model scans sessions and selects high-attention patches as compact evidence units. In \textbf{(b) Index Construction}, these patches are contextualized at the session level by an LLM; each indexed entry stores both a patch embedding and a session-aware patch summary. In \textbf{(c) Retrieval-Augmented LLM Reasoning}, query patches are paired with their most similar indexed counterparts, giving the LLM cross-session references for joint reasoning. It evaluates patches within their session context and retrieved signals, producing patch-level risk scores, saliency estimates, and a coherent session-level assessment. In \textbf{(d) Cross-Granularity Distillation}, the LLM outputs supervise the lightweight model: patch risk and saliency are aligned with session-level judgments, transferring the reasoning chain from local cues to overall risk and strengthening interpretability. Finally, in \textbf{(e) Inference}, PatchNet alone yields efficient session-level predictions while retaining internal patch-level evidence as explanatory cues.
Figure~\ref{fig:overview} presents the overall framework of \textbf{CS-VAR} for live streaming risk assessment. The framework follows a coarse-to-fine paradigm where a domain-tailored lightweight model~(PatchNet) and a retrieval-augmented LLM collaborate in complementary roles. 

First, \textbf{(a) PatchNet Warm-up} stage scans incoming training sessions and highlights high-attention patches as compact evidence units. These patches are then contextualized at the session level. In \textbf{(b) Index Construction}, an LLM generates session-aware summaries for key patches and stores them alongside patch embeddings. During \textbf{(c) Retrieval-Augmented LLM Reasoning}, query patches are paired with their most similar indexed counterparts, providing cross-session references that support joint reasoning. The LLM evaluates each patch within both its session context and retrieved signals, yielding patch-level risk scores, saliency estimates, and an integrated session-level assessment. Building on this, \textbf{(d) Cross-Granularity Distillation} distills LLM reasoning into the small model, linking patch-level cues to session-level risk and producing interpretable, actionable signals. Finally, in \textbf{(e) Inference}, PatchNet alone performs efficient session-level predictions while retaining patch-level evidence as explanatory cues.

\subsection{Small Model Backbone: PatchNet}

We design a lightweight backbone, \textbf{PatchNet}, tailored for the live streaming scenario where risks often unfold as fragmented cues and chain-like patterns across time and users.  
Its warm-up stage serves two purposes:
(1) learning patch embeddings that capture localized risk cues within each session, and
(2) aligning these embeddings for cross-session retrieval and augmentation.

\noindent
\textbf{Patch Encoder.}  
To preserve both global context and local dynamics in live streaming sessions, we design a two-stage encoder. Given a session as a time-ordered action sequence $S=[\alpha_1,\alpha_2,\dots,\alpha_{N_s}]$ with actions $\alpha_i=(u_i,t_i,a_i,x_i)$, each action is first embedded as
$\mathbf{e}_i = \big[\,\mathbf{e}_{a_i} \;\|\; \mathrm{Proj}(x_i)\,\big],$
where $\mathbf{e}_{a_i}$ is a learnable action-type embedding and $\mathrm{Proj}(\cdot)$ maps the pre-encoded textual content $x_i$ into the latent dimension $d_k$.

The flattened action sequence $[\mathbf{e}_1,\dots,\mathbf{e}_{N_s}]$ is processed by a Transformer encoder~\cite{vaswani2017attention} to capture long-range inter-action dependencies, producing contextualized tokens $[\mathbf{h}_1,\dots,\mathbf{h}_{N_s}]$. We then group these contextualized tokens into user–timeslot patches: a patch $P_{u,k}$ collects the action tokens of user $u$ within timeslot $\mathcal{T}_k$: $p_{u,k}=\{\mathbf{h}_i \mid t_i\in\mathcal{T}_k,\; u_i=u\}
$. Each token sequence within a patch is then encoded by an LSTM, and the final hidden state serves as the patch embedding:
\begin{equation}
\mathbf{p}_{u,k}=\mathrm{LSTM}\bigl([\mathbf{h}_i]_{\alpha_i\in p_{u,k}}\bigr)[-1] \in \mathbb{R}^{d_k}.
\end{equation}

\noindent
\textbf{Relational Graph Construction.}  
Live-streaming risks frequently arise from structured dependencies: bursty actions in neighboring time windows, recurring or evolving risky behaviors by the same user, and host–audience interactions. To model these patterns, we build relation-specific adjacency matrices $\{\mathbf{A}^{(\zeta)}\}_{\zeta\in\mathcal{R}}$ over all patches, with $\mathcal{R}=\{\mathrm{t},\mathrm{u},\mathrm{r},\mathrm{a}\}$
 for temporal, user, role-guided, and auxiliary relations. Each relation matrix is formed by combining structural connectivity with semantic affinity:
\begin{equation}
\begin{aligned}
\mathbf{A}^{\mathrm{t}}_{ij} &= \mathbf{1}[\,|t_i - t_j| \le 1\,] \cdot \mathrm{sim}(\mathbf{p}_i,\mathbf{p}_j), \quad
\mathbf{A}^{\mathrm{u}}_{ij} = \mathbf{1}[\,u_i = u_j\,] \cdot \mathrm{sim}(\mathbf{p}_i,\mathbf{p}_j), \\
\mathbf{A}^{\mathrm{r}}_{ij} &= \mathbf{1}[\, (u_i = u^{\texttt{h}}, u_j \neq u^{\texttt{h}}) \lor (u_j = u^{\texttt{h}}, u_i \neq u^{\texttt{h}})\,] \cdot \mathrm{sim}(\mathbf{p}_i,\mathbf{p}_j), \\
\mathbf{A}^{\mathrm{a}}_{ij} &= \mathrm{sim}(\mathbf{p}_i,\mathbf{p}_j) - \max\!\left(\mathbf{A}^{\mathrm{t}}_{ij}, \mathbf{A}^{\mathrm{u}}_{ij}, \mathbf{A}^{\mathrm{r}}_{ij}\right),
\end{aligned}
\end{equation}
where $\mathrm{sim}(\mathbf{p}_i,\mathbf{p}_j)$ is a normalized dot-product similarity, $t_i$ and $u_i$ are patch $i$'s timeslot and user, and $u^{\texttt{h}}$ denotes the host.  
These adjacency matrices respectively capture bursty temporal coordination, evolving intent from the same user, host–audience interplay, and residual semantic affinity beyond structured links.

The relation-aware adjacency matrices are combined adaptively, as $\gamma^{(\zeta)}$ controls the relative importance of each relation:
\begin{equation}
\mathbf{A}^{\mathrm{G}}=\sum_{\zeta\in\mathcal{R}}\gamma^{(\zeta)}\cdot\mathbf{A}^{(\zeta)},\qquad
\mathbf{A}^{\mathrm{G}}_{ij}=\frac{\exp(\mathbf{A}^{\mathrm{G}}_{ij})}{\sum_{j'}\exp(\mathbf{A}^{\mathrm{G}}_{ij'})}.
\end{equation}

\noindent
\textbf{Graph-Aware Patch Relational Learning.}  
Subsequently, patch embeddings $\{\mathbf{p}_i\}$ are refined by a Graph-Aware Transformer~\cite{shehzad2024graph} whose self-attention is biased by the relation-aware adjacency $\mathbf{A}^{\mathrm{G}}$:
\begin{equation}
\mathrm{Attention}(\mathbf{Q},\mathbf{K},\mathbf{V},\mathbf{A}^{\mathrm{G}})=
\mathrm{Softmax}\!\left(\frac{\mathbf{Q}\mathbf{K}^\top+\mathbf{A}^{\mathrm{G}}}{\sqrt{d_k}}\right)\mathbf{V}.
\label{eq:att score}
\end{equation}
As the Transformer input, we flatten the patch grid in user-then-timeslot order, which preserves user continuity while maintaining temporal progression, and prepend a learnable $[\texttt{CLS}]$ token to this ordered sequence. The learned hidden state of $[\texttt{CLS}]$ is taken as the session embedding $\mathbf{H}_s$. Then, a session head (implemented as a 2-layer MLP) maps $\mathbf{H}_s$ to the session risk prediction $\hat{y}_s$.
This design lets patches interact under both semantic affinity and structural priors while producing a compact session representation for prediction.

\noindent
\textbf{Learning Objective of Warm-Up.} During warm-up, PatchNet is optimized with a session-level binary cross-entropy~(BCE) loss over training sessions:
\begin{equation}
\mathcal{L}_{\mathrm{s}} = - \sum_{s=1}^N \Bigl[ y_s \log(\hat{y}_s) + (1 - y_s) \log(1 - \hat{y}_s) \Bigr],
\label{eq:loss:session bce loss}
\end{equation}
where $y_s\in\{0,1\}$ is the session label and $\hat{y}_s$ is produced by the session head above. Note that we warm up PatchNet until validation performance starts to decline, and use the corresponding checkpoint for subsequent stages.
Optimized for the session-level objective, the refined patch embeddings function as retrieval-ready vectors for cross-session indexing. Furthermore, the learned cross-patch attention provides a natural criterion for identifying key patches within a session.

\subsection{Index Construction}
\label{sec:met:index}
Building on the PatchNet warm-up stage, we identify key patches using session-level cross-patch attention and construct a retrieval index that integrates refined patch embeddings with LLM-generated interpretations contextualized at the session level.

\noindent
\textbf{Key Patch Selection.}  
With reference to Eq.~\ref{eq:att score}, we treat the $[\texttt{CLS}]$ token as the query and extract its attention distribution over patches:
\begin{equation}
\boldsymbol{\alpha}^{[\texttt{CLS}]} = \mathrm{Softmax}\!\left( 
\frac{ \mathbf{Q}_{\texttt{[CLS]}} \mathbf{K}^\top + \mathbf{A}^{\mathrm{G}}_{\texttt{[CLS]}, :} }{ \sqrt{d_k} } 
\right).
\end{equation}
Each entry $\boldsymbol{\alpha}^{[\texttt{CLS}]}_j$ measures the contribution of patch $j$ to the session-level representation. Patch selection is performed only for sessions predicted as positive by PatchNet, since in such sessions, the highly attended patches are more likely to highlight fragments that are already associated with potential risk by the small model.

To provide a comprehensive view, we select patches from the host and viewers separately. Since the host primarily drives the narrative and risk exposure of the session, we retain the top-5 host patches to capture diverse yet central cues. In contrast, viewer behavior is fragmented and noisy; we keep only the top-3 viewer patches to ensure diversity while mitigating spurious signals.

\noindent
\textbf{Indexing and Semantic Augmentation.}  
Key patches are then organized and processed at the session level rather than in isolation. Specifically, each patch is first represented by concatenating its raw action texts, which are often fragmented when viewed alone. All such patch segments from the same session are then jointly provided to the LLM, which abstracts each into a concise session-aware summary. Instead of plain action concatenation, this design produces context-sensitive interpretations that capture not only surface events but also their dependencies within the broader interaction chain, thereby exposing underlying user intentions. 

Compared to conventional RAG pipelines that directly index verbatim text chunks~\cite{lewis2020retrieval,fan2024survey,gao2023retrieval}, our index construction leverages risk-aware embeddings from PatchNet and session-contextualized LLM summaries, yielding entries that are both retrieval-ready and semantically enriched. A toy example of the prompting format and response is shown below (see Appendix~\ref{sec:app:prompt} for a detailed version).

\begin{tcolorbox}[
  colframe=purple,
  width=\linewidth,
  arc=0.5mm,
  auto outer arc,
  title={Session-Aware Summarization Toy Prompt \& Response},
  boxsep=2pt,        % 内边距整体缩小
  left=2pt,
  right=2pt,
  top=2pt,
  bottom=2pt,
  fontupper=\small   % 可选：文字稍微小一点，更紧凑
]
A live streaming session has: [Patch 1, Patch 2, Patch 3]

Q: Summarize each patch considering the whole session and potential risk chains.

A:
[\{"patch\_id": "1", "summary": "Summary of Patch 1"\}, ...]
\end{tcolorbox}

\noindent
Formally, each indexed item is stored in Patch Index $\mathcal{I}_p$ as $
\text{entry}_j = \big(\mathbf{p}_j,\tau_j,\text{meta}_j\big), j \in \mathcal{I}_p,$
where $\mathbf{p}_j$ is the normalized patch embedding from PatchNet for similarity search, $\tau_j$ is the LLM-generated session-aware summary, and $\text{meta}_j$ includes identifiers such as session id. The resulting entries are stored in a FAISS index~\cite{johnson2019billion} with cosine similarity, enabling efficient cross-session retrieval for LLM reasoning.

\subsection{Retrieval-Augmented LLM Reasoning}
\label{sec:met:llm}

While PatchNet highlights risk-sensitive patches within a session, its reasoning remains confined to intra-session cues. However, many behaviors are benign in isolation and only become indicative when similar patterns recur across different sessions. Capturing such ``déjà vu'' repetitions is crucial for revealing latent risk trajectories. To this end, we introduce a retrieval-augmented LLM reasoning stage that complements local evidence with cross-session context.
\noindent
\textbf{Query Formation.}  
For each training session \(s\), we apply PatchNet’s cross-patch attention to identify the most risk-relevant fragments. We retain the top-5 host patches and top-3 viewer patches, forming the key patch set \(\mathcal{P}^{\texttt{key}}_s\) with at most eight elements. Each patch \(k \in \mathcal{P}^{\texttt{key}}_s\) is represented by its normalized embedding \(\mathbf{p}_k\), which directly serves as a query vector for retrieval. This patch-level design naturally aligns queries and index entries in the same embedding space, since both are derived from PatchNet. The use of PatchNet embeddings injects a risk-sensitive inductive bias into retrieval, unlike conventional RAG that relies on generic encoders.

\noindent
\textbf{Cross-Session Retrieval.}  
Given a query embedding \(\mathbf{p}_k\), we perform similarity search over the patch index \(\mathcal{I}_p\) using cosine distance, while excluding candidates from the same session to avoid trivial matches:
\begin{equation}
\mathcal{N}_k = \operatorname*{arg\,max}_K 
\big\{ \cos(\mathbf{p}_k, \mathbf{p}_{j'}) : j' \in \mathcal{I}_p,\, s(j') \neq s(k) \big\}.
\end{equation}

Each retrieved patch $\text{entry}_j'=(\mathbf{p}_{j'}, \tau_{j'}, \text{meta}_{j'})$ provides its embedding and the session-aware summary \(\tau_{j'}\), with \(s(j')\) denoting the session identifier stored in metadata. In practice, we set \(K=1\) per query, so a session yields at most eight neighbors. These neighbors are then bundled with the original session’s patches to form the evidence package for LLM reasoning.

Unlike conventional RAG pipelines that rely on generic encoders, CS-VAR's retrieval operates entirely in the embedding space of PatchNet. This design injects PatchNet’s risk-sensitive inductive bias into the retrieval process: neighbors are selected not only for semantic similarity, but for their alignment with the small model’s discriminative focus. Together with the session-aware summaries attached to each entry, the retrieved evidence is both interpretable and tailored for downstream LLM reasoning.

\noindent
\textbf{LLM Reasoning with Retrieved Evidence.}  
While retrieval supplies cross-session neighbors, their role is not to stand alone but to serve as auxiliary evidence when reasoning over the query session. To this end, we design an Evidence-Integrated Reasoning Prompt that presents both local patches and their retrieved neighbors, guiding the LLM to output structured predictions across granularities.

Each prompt asks the LLM to assign (i) patch-level risk scores and saliency weights, indicating the riskiness and contribution of each fragment, and (ii) a session-level risk score that synthesizes local cues with cross-session context. A minimal toy example is shown below, with a detailed version presented in Appendix~\ref{sec:app:prompt}:
\begin{tcolorbox}[
  colframe=purple,
  width=\linewidth,
  arc=0.5mm,
  auto outer arc,
title={Evidence-Integrated Reasoning Toy Prompt \& Response},
  boxsep=2pt,        % 内边距整体缩小
  left=2pt,
  right=2pt,
  top=2pt,
  bottom=2pt,
  fontupper=\small   % 可选：文字稍微小一点，更紧凑
]
A query session has patches paired with retrieved neighbors:  
[ (Patch 1, Retrieved 1), (Patch 2, Retrieved 2), ... ]

Q: Assess each patch considering both the query and its retrieved neighbor.  
Output patch-level risk scores and saliency weights, and an overall session risk.

A: [
  \{"patch\_id": "1", "risk\_score": 0.7, "saliency": 0.8\}, 
  \{"patch\_id": "2", "risk\_score": 0.1, "saliency": 0.3\}, 
  ...,
  \{"session\_risk": 0.75\}
]
\end{tcolorbox}

\noindent
Formally, for a query session \(s\) with key patch set \(\mathcal{P}^{\texttt{key}}_s\), the LLM outputs
\[
\big\{ \big(y^{\texttt{LLM}}_k, \texttt{SAL}_k\big) : k \in \mathcal{P}^{\texttt{key}}_s \big\}, 
\quad {y}_s^{\texttt{LLM}},
\]
where \(y^{\texttt{LLM}}_k \in [0,1]\) is the \emph{patch-level risk score}, 
\(\texttt{SAL}_k \in [0,1]\) is the \emph{patch-level saliency}, 
and \({y}_s^{\texttt{LLM}} \in [0,1]\) is the \emph{session-level risk score}.

This structured reasoning grounds session predictions in concrete patch-level evidence and explicitly links local cues to global assessment, providing supervision signals for distilling LLM reasoning into PatchNet.

\subsection{Cross-Granularity Distillation}

Building on the warm-up checkpoint, we further optimize PatchNet with supervision from LLM reasoning outputs. This stage completes our full framework, CS-VAR, by distilling the LLM’s cross-granularity judgments into the small model. The goal is to make PatchNet imitate the fine-grained prediction and the reasoning chain while retaining efficiency for real-time inference.

Recall that $\hat{y}_s$ denotes PatchNet’s predicted risk score for session $s$, and $y_s \in \{0,1\}$ the ground-truth label. For each key patch $k \in \mathcal{P}^{\texttt{key}}_s$, PatchNet produces a prediction $\hat{y}_k$ through a 2-layer MLP patch head, while the LLM provides a patch-level risk score $y^{\texttt{LLM}}_k$, a saliency weight $\texttt{SAL}_k$, and a session-level score ${y}^{\texttt{LLM}}_s$. The overall objective integrates three components, with $\beta$ and $\delta$ controlling the weights of the auxiliary terms:
\begin{equation}
\mathcal{L} = \mathcal{L}_{\text{s}}
+ \beta \,\mathcal{L}_{\text{p}}
+ \delta \,\mathcal{L}_{\text{p2s}}.
\label{eq:loss:whole}
\end{equation}

\noindent
\textbf{Session loss.}  
$\mathcal{L}_{\text{s}}$ is a binary cross-entropy between $\hat{y}_s$ and $y_s$ ~(see Eq.~\ref{eq:loss:session bce loss}), anchoring PatchNet to ground-truth labels as in warm-up.

\noindent
\textbf{Patch loss.}  
$\mathcal{L}_{\text{p}}$ enforces fine-grained alignment at the patch level:
\begin{equation}
\mathcal{L}_{\text{p}} = 
\frac{1}{|\mathcal{P}^{\texttt{key}}_s|}
\sum_{k \in \mathcal{P}^{\texttt{key}}_s}
\big(\hat{y}_k - y^{\texttt{LLM}}_k\big)^2,
\label{eq:patch loss}
\end{equation}
where $\hat{y}_k \in [0,1]$ is PatchNet’s predicted risk score for patch $k$.

\noindent
\textbf{Patch-to-session loss.}  
Since the LLM’s prediction $\hat{y}^{\texttt{LLM}}_s$ is based only on the retrieved \emph{key patches} rather than the full session, it reflects a limited view of the session. We require PatchNet to reproduce this partial perspective by weighting its own patch predictions with the LLM’s saliency:
\begin{equation}
\tilde{y}_s =
\sum_{k \in \mathcal{P}^{\texttt{key}}_s}
\frac{\texttt{SAL}_k}{\sum_{j \in \mathcal{P}^{\texttt{key}}_s}\texttt{SAL}_j}
\, \hat{y}_k,
\qquad
\mathcal{L}_{\text{p2s}} = (\tilde{y}_s - y^{\texttt{LLM}}_s)^2.
\end{equation}
This loss ensures that PatchNet’s patch-level predictions, when aggregated through the LLM’s notion of saliency, remain consistent with the LLM’s session-level output under the same partial view.

Together, the three terms capture a hierarchy of supervision:  
(1) $\mathcal{L}_{\text{p}}$ distills \emph{fine-grained judgments} at the patch level,  
(2) $\mathcal{L}_{\text{p2s}}$ enforces consistency in the \emph{key-patch local view}, and  
(3) $\mathcal{L}_{\text{s}}$ grounds PatchNet in \emph{global labels}.  
This cross-granularity design explicitly transfers the reasoning chain from LLM to PatchNet, bridging local cues and global decisions, and thus equips CS-VAR with the ability to deliver interpretable and real-time risk assessment.

\subsection{Training and Inference}

\noindent
\textbf{Training.}  
CS-VAR is trained in four stages:
(1) \emph{Warm-up:} PatchNet is trained with a session-level BCE loss (Eq.~\ref{eq:loss:session bce loss}) to learn retrieval-ready patch embeddings and to expose risk-sensitive cues.  
(2) \emph{Index construction:} the learned embeddings and LLM summaries are organized into a cross-session patch index.  
(3) \emph{Retrieval-augmented reasoning:} the LLM consumes key patches together with their retrieved neighbors and produces structured multi-granularity judgments.  
(4) \emph{Cross-granularity distillation:} PatchNet is further optimized with a multi-task objective (Eq.~\ref{eq:loss:whole}) that aligns its predictions with both ground-truth labels and the LLM’s reasoning outputs.

\noindent
\textbf{Inference.}  
At test time, CS-VAR operates solely with PatchNet, producing both session-level risk predictions and fine-grained signals that enhance interpretability for moderation. By distilling the LLM’s cross-session reasoning, the model captures recurring cues and risk-chain patterns while maintaining efficiency.

 \section{Experiments}

We evaluate CS-VAR through offline experiments and online deployment, aiming to answer the following research questions:

\begin{itemize}[leftmargin=*,topsep=5pt]
    \item \textbf{RQ1:} How does CS-VAR perform in live streaming risk assessment compared to baselines?
    \item \textbf{RQ2:} What is the contribution of each individual component in the CS-VAR pipeline?
    \item \textbf{RQ3:} How does patch-level evidence reveal actionable patterns of risk in live streaming?
    \item \textbf{RQ4:} Does CS-VAR produce representations that capture recurring scripted schemes across live streaming sessions?
    \item \textbf{RQ5:} How effective is CS-VAR in online deployment?
\end{itemize}

\subsection{Experimental Setup}
\noindent
\textbf{Datasets.}
\begin{table}[b]
\caption{Statistics of the May and June datasets.}
\label{table:exp:dataset}
\centering
\resizebox{0.95\linewidth}{!}{
  \begin{tabular}{c|c|cccc} 
  \toprule
   &  & \#sessions & \#Avg.actions & \#Avg.users & Avg.time~(min) \\ 
  \midrule
  \multirow{3}{*}{\textbf{May}} & train & $176,347$ &709  & 35 & 30.0 \\ 
   & val & $23,562$ & 704 & 36 & 29.6 \\ 
   & test & $22,462$ & 740 & 37 & 29.7 \\ 
  \midrule
  \multirow{3}{*}{\textbf{June}} & train & $79,552$ & 700 & 36 & 30.0 \\ 
   & val & $10,934$ & 767 & 40 & 29.1 \\ 
   & test & $10,967$ &725 & 37 &29.1  \\ 
  \bottomrule
  \end{tabular}
}
\end{table}
For offline evaluation, we construct two large-scale real-world datasets from a major live streaming platform\footnote{All data were collected in compliance with the platform’s privacy policy.}, denoted as \textbf{May} and \textbf{June}. 
Table~\ref{table:exp:dataset} reports the basic statistics of the \textbf{May} and \textbf{June} datasets. 
\textbf{May} contains training data from 05/20/2025 to 06/03/2025, validation data from 06/11/2025 to 06/12/2025, and test data from 06/13/2025 to 06/14/2025. 
\textbf{June} includes training data from 06/04/2025 to 06/10/2025, validation data on 06/15/2025, and test data on 06/16/2025.
%Each dataset consists of action sequences truncated to the first 30 minutes to support early-stage risk detection. 
%All positive cases are preserved, while negatives are subsampled at a 1:10 ratio, resulting in a positive proportion of about 9\%. 

\textit{Preprocessing.} Data construction follows three main steps: 
(i) truncate each session to the first 30 minutes and discretize into 100-second slots, 
(ii) subsample negative sessions at a 1:10 ratio, 
and (iii) filter out inactive users, and select the 50 most active viewers per session to form the action sequence.

\textit{Action Space.} Viewer activities include session entry, comments (including danmaku), gifting, likes, shares, leaderboard appearances, group joins, and co-stream requests. 
host actions cover stream initiation, speech transcripts generated by voice-to-text models, and visual content extracted via OCR. 
The latter two are produced by scheduled inspection mechanisms, where logs are discretized during collection as part of the platform’s risk control workflow.

\textit{Encoding.} Sequences are capped at 2,096 tokens. 
Textual fields such as comments and spoken transcripts are pre-encoded with a Chinese-BERT model\footnote{https://huggingface.co/google-bert/bert-base-chinese}. 
All interaction data is sourced from publicly available content and processed in strict accordance with the platform’s privacy and security policies.

\noindent
\textbf{Baselines.}
Since there is limited prior work addressing session-level risk assessment in live streaming, we compare CS-VAR against two groups of baselines. 
(i) \textit{Sequence models} that directly capture temporal dependencies in the action sequences of lives: \textbf{Transformer}~\cite{vaswani2017attention}, \textbf{Reformer}~\cite{kitaev2020reformer}, and \textbf{Informer}~\cite{zhou2021informer}. 
(ii) \textit{Instance aggregation methods}, originally proposed for multiple instance learning~(MIL), are relevant to our task, as both settings require aggregating local signals into a global decision. In our case, session-level predictions are made by combining signals from multiple patches within a session: \textbf{mi-NET}~\cite{wang2018revisiting}, \textbf{Attention MIL~(AtMIL)}~\cite{ilse2018attention}, \textbf{Additive MIL~(AdMIL)}~\cite{javed2022additive}, \textbf{MIL-LET}~\cite{early2024inherently}, \textbf{TimeMIL}~\cite{chen2024timemil}, and \textbf{TAIL-MIL}~\cite{jang2025tail}.
More details are provided in App.~\ref{app:baselines}.

\noindent
\textbf{Implementation Details.}
All offline experiments are conducted using Python 3.11.2 and PyTorch 2.6.0. 
We employ the \textit{doubao-1.5-pro-32k}\footnote{https://www.volcengine.com/docs/82379/1554678} API for LLM reasoning.
We train all models for up to 100 epochs using the AdamW optimizer with a learning rate of $1\mathrm{e}{-4}$ and a weight decay of $1\mathrm{e}{-4}$. 
The batch size is fixed at 128. 
Early stopping is applied with a patience of 20 epochs to mitigate overfitting. 
We use a dropout rate of 0.1 and set the embedding dimension to 128 for all methods. 
For Transformer-based architectures, the number of attention heads is set to 8. 
Sequence encoders (including LSTM, and the Transformer modules in both baselines and CS-VAR) use two layers, whereas the Graph-Aware Transformer uses a single layer. 
For the composite loss, the coefficients $\beta$ and $\delta$ are selected via grid search over $\{0.5, 1.0, 1.5, 2.0\}$.

\noindent
\textbf{Evaluation Metrics.}
%We report \textbf{PR-AUC}, \textbf{F1-score}, \textbf{R@0.1FPR}, and \textbf{FPR@0.9R}. 
Our evaluation employs the following four metrics: \textbf{PR-AUC}, \textbf{F1-score}, \textbf{R@0.1FPR}, and \textbf{FPR@0.9R}. 
PR-AUC and F1-score characterize the precision–recall trade-off, which is critical in imbalanced classification. 
ROC-AUC, though widely used, may overestimate performance when negatives dominate; PR-AUC instead emphasizes the model’s capacity to recover positive cases. 
R@0.1FPR measures recall when the false positive rate is limited to 10\%, while FPR@0.9R quantifies the false alarm rate under 90\% recall. 

%\noindent
%Note that additional descriptions of datasets, baselines, training configurations, and metric definitions are provided in Appendix \ref{sec:app:exp}.

\subsection{Overall Performance~(RQ1)}
\begin{table*}[t]
\centering
\caption{Performance comparison of CS-VAR with baselines. The best and second-best results among all methods are shown in bold and underline, respectively. A `*' marks statistically significant gains over the best-performing baseline (p-value $<$ 0.05).
}
\label{tab:exp:main}

% 用 resizebox 缩放到列宽（\columnwidth），高度按比例自动调整
\resizebox{\textwidth}{!}{
\begin{tabular}{@{}c|c|cccc|cccc@{}}   % 分类列|方法列|4指标|4指标
\toprule
\multicolumn{2}{@{}c|}{\multirow{3}{*}{\textbf{Methods}}} & \multicolumn{4}{c|}{\textbf{May Dataset}} & \multicolumn{4}{c@{}}{\textbf{June Dataset}} \\ 
\cmidrule(lr){3-6} \cmidrule(lr){7-10}
\multicolumn{2}{@{}c|}{} & \makecell{PR-AUC $\uparrow$} & F1-score $\uparrow$ & R@0.1FPR $\uparrow$ & FPR@0.9R $\downarrow$
& PR-AUC $\uparrow$ & F1-score $\uparrow$ & R@0.1FPR $\uparrow$ & FPR@0.9R $\downarrow$ \\ 
\midrule
% Sequence Models 分类（合并3行，用 shortstack 换行）
\multirow{3}{*}{{\shortstack{\textit{Sequence}\\ \textit{Models}}}}
& Transformer   & 0.7189 & 0.6668 & 0.8394 & 0.1580  & 0.6801 & 0.6341  & 0.8225 & 0.1565 \\
& Reformer     & 0.7203 & 0.6752 & 0.8575 & 0.1436 & 0.6911  & 0.6395 & 0.8104 & 0.1760 \\
& Informer   & 0.7246 & 0.6708 & 0.8438 & 0.1555 & 0.6879 & 0.6391  & 0.8375 & 0.1601 \\
\midrule
% MIL methods 分类（合并6行，用 shortstack 换行）
\multirow{6}{*}{{\shortstack{\textit{Instance}\\ \textit{Aggregation}\\ \textit{Methods}}}}
& mi-NET    & 0.7276 & 0.6769 & 0.8560  & \underline{0.1320} & 0.6911 & 0.6406 & 0.8225 & 0.1673  \\
& AtMIL    & 0.7252 & 0.6763 & 0.8550 & 0.1441 & 0.6952 & 0.6519  & 0.8415 & 0.1479   \\
& AdMIL     & 0.7196 & 0.6781 & 0.8389 & 0.1568 & 0.6837 & 0.6491 & 0.8225 & 0.1565 \\
& MIL-LET   & 0.7241 & 0.6749 & 0.8546 & 0.1418 & 0.6942 & \underline{0.6528} & 0.8455 & 0.1499  \\
& TimeMIL   & \underline{0.7353} & \underline{0.6790} & \underline{0.8599} & 0.1436  & 0.6963 & 0.6471  & \underline{0.8495} & \underline{0.1367} \\
& TAIL-MIL   & 0.7316 & 0.6785 & 0.8570 & 0.1341  & \underline{0.7029} & 0.6509 & 0.8205 & 0.1555 \\
\midrule
\multicolumn{2}{@{}c|}{\textbf{CS-VAR}(\textit{Ours})}   & \textbf{0.7721*} & \textbf{0.7099*} & \textbf{0.8761*} & \textbf{0.1190*} & \textbf{0.7322*} & \textbf{0.6822*} & \textbf{0.8636*} & \textbf{0.1338*} \\
\multicolumn{2}{@{}c|}{\textbf{Best Improv.}} &
\textit{+5.0\%} & \textit{+4.6\%} & \textit{+1.9\%} & \textit{-10.9\%} & 
\textit{+4.2\%} & \textit{+4.5\%} & \textit{+1.7\%} & \textit{-2.2\%} \\ 
\bottomrule
\end{tabular}
}  % 关闭 resizebox
\end{table*}
To address \textbf{RQ1}, we compare CS-VAR with strong baselines on both the May and June datasets. 
The results are summarized in Table~\ref{tab:exp:main}, from which we draw the following observations.

First, CS-VAR consistently achieves the best performance across all metrics and both datasets. 
The improvements over the strongest baseline are substantial: for example, on the May dataset, CS-VAR raises PR-AUC by 5.0\% and reduces FPR@0.9R by 10.9\%.

Second, among baselines, instance aggregation methods (e.g., TimeMIL and TAIL-MIL) generally outperform sequence models such as Transformer and Reformer. 
This indicates that aggregating local signals into session-level decisions is more effective than modeling long sequences directly. 
However, even the strongest MIL-based methods remain behind CS-VAR, suggesting that our cross-session evidence retrieval and patch-to-session design captures richer dependencies that traditional MIL pooling cannot.

Third, the performance trends are consistent between the May and June datasets. 
In both cases, CS-VAR not only improves recall at low false positive rates (R@0.1FPR) but also lowers false alarm rates when recalling 90\% of risky cases (FPR@0.9R). 
This demonstrates that CS-VAR balances precision and recall under the strict requirements of risk control, and generalizes well across different time spans of traffic.

Overall, these results validate the effectiveness of CS-VAR in live streaming risk assessment, showing clear advantages over both sequence models and MIL-style instance aggregation methods.

\begin{table}[b]
%\vspace{-5mm}
\centering
\caption{Ablation study of CS-VAR on the May dataset.}
\label{tab:exp:ablation}
\resizebox{0.95\columnwidth}{!}{
  \begin{tabular}{@{}c|cccc@{}} 
  \toprule
  \textbf{} & \textbf{PR-AUC $\uparrow$} & \textbf{F1-score $\uparrow$} & \textbf{R@0.1FPR $\uparrow$} & \textbf{FPR@0.9R $\downarrow$} \\ 
  \midrule
  CS-VAR & \textbf{0.7721} & \textbf{0.7099} & \textbf{0.8761} & \textbf{0.1190} \\
  \midrule
  CS-VAR \textit{w/o G} & 0.7642 & 0.7027 & 0.8746 & 0.1253 \\
  CS-VAR \textit{w/o R} & 0.7616 & 0.7023 & 0.8639 & 0.1318 \\
  CS-VAR \textit{w/o L} & 0.7614 & 0.6948 & 0.8737 & 0.1290 \\
  CS-VAR \textit{w/o D} & 0.7592 & 0.6930 & 0.8712 & 0.1262 \\
  \bottomrule
  \end{tabular}
}
\end{table}
\subsection{Ablation Study~(RQ2)}

To address \textbf{RQ2}, we perform ablations on the May dataset (Table~\ref{tab:exp:ablation}) using the following variants, each targeting a specific stage:

\noindent \textbf{\textit{(1)~w/o G}}: Replacing the graph-aware attention in PatchNet with vanilla self-attention consistently degrades performance, showing that structured attention is necessary to capture cross-entity dependencies beyond plain sequences.  

\noindent  \textbf{\textit{(2)~w/o R}}: Removing the retrieval index leaves LLM reasoning with only within-session patches, lowering performance for limiting the detection of recurring scripted behaviors.  

\noindent   \textbf{\textit{(3)~w/o L}}: Replacing LLM reasoning with a simple retrieval-augmented tactic, where the query patch is averaged with its top-1 nearest neighbor embedding, further degrades performance. This suggests that LLM structured reasoning contributes far more than raw similarity. 

\noindent \textbf{\textit{(4)~w/o D}}: Training PatchNet only with session-level supervision yields the steepest degradation, as the model can no longer absorb cross-session reasoning distilled from the LLM. 
\iffalse
\begin{itemize}[leftmargin=*,topsep=5pt]
\item
    \textbf{\textit{w/o G}}: Replacing the graph-aware attention in PatchNet with vanilla self-attention consistently degrades performance, showing that structured attention is necessary to capture cross-entity dependencies beyond plain sequences.  
    
    \item \textbf{\textit{w/o R}}: Removing the retrieval index leaves LLM reasoning with only within-session patches, lowering performance for limiting the detection of recurring scripted behaviors.  
    
    \item \textbf{\textit{w/o L}}: Replacing LLM reasoning with a simple retrieval-augmented tactic, where the query patch is averaged with its top-1 nearest neighbor embedding, further undergrades performance. This suggests that LLM structured reasoning contributes far more than raw similarity. 
    
    \item \textbf{\textit{w/o D}}: Training PatchNet only with session-level supervision yields the steepest degradation, as the model can no longer absorb cross-session reasoning distilled from the LLM. 
    %In effect, distillation is what carries the benefits of retrieval-augmented LLM reasoning into the compact backbone.  
\end{itemize}
\fi

Taken together, the results suggest that the four stages reinforce each other, and it is their integration that enables CS-VAR to effectively transfer cross-session reasoning into the backbone and thereby strengthen risk control in live streaming.

\subsection{Case Study~(RQ3)}
\begin{figure}[b] \centering \includegraphics[width=0.48\textwidth]{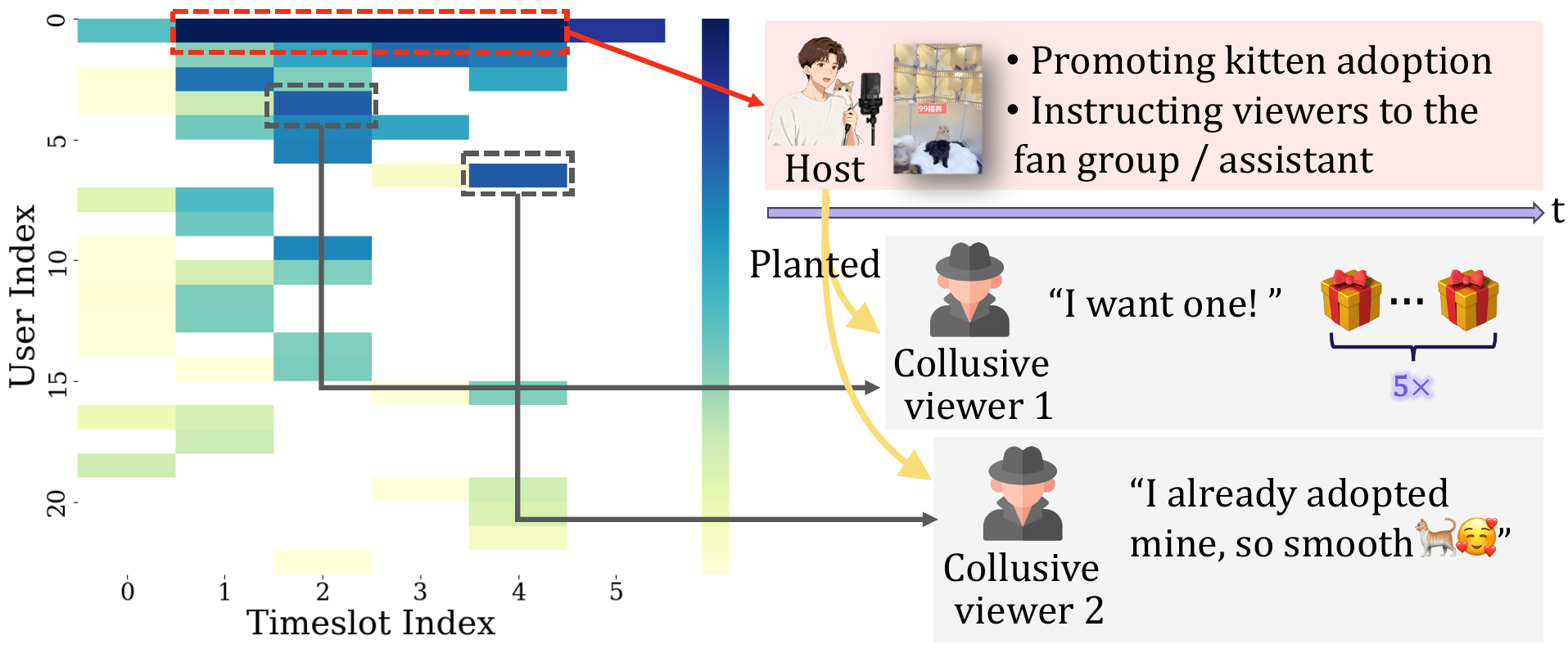} 
\caption{
A case of a kitten adoption scam detected by CS-VAR. 
\textbf{Left:} User–timeslot patch grid heatmap showing risk scores across the session. Scores from CS-VAR's patch head as in Eq.~\ref{eq:patch loss}.
\textbf{Right:} The host promotes low-cost adoption and fan-group/assistant contact. Two viewers coordinate: one expresses interest then sends small gifts, another posts a testimonial claiming successful adoption.
}
\label{fig:exp:case_cat} \end{figure}
To address \textbf{RQ3}, we present a case study highlighting coordinated host–viewer behaviors. As shown in Figure~\ref{fig:exp:case_cat}, CS-VAR identifies segments where promotional messages align with staged viewer responses, revealing recurring scripted patterns.

Here, the host advertises a \emph{low-cost kitten adoption}. One viewer repeatedly sends small gifts, boosting engagement, while another posts a nearly identical testimonial claiming a successful adoption. The host also urges the audience to join the fan group and contact an assistant for specific breeds. Although different from typical schemes such as cheap phone sales and part-time job scams, the pattern of staged persuasion and engineered engagement is captured by CS-VAR.
This demonstrates how CS-VAR surfaces interpretable, cross-session cues to support evidence-based risk moderation.

\subsection{Representation Analysis (RQ4)}

\begin{figure}[t]
    \centering
    \includegraphics[width=0.9\linewidth]{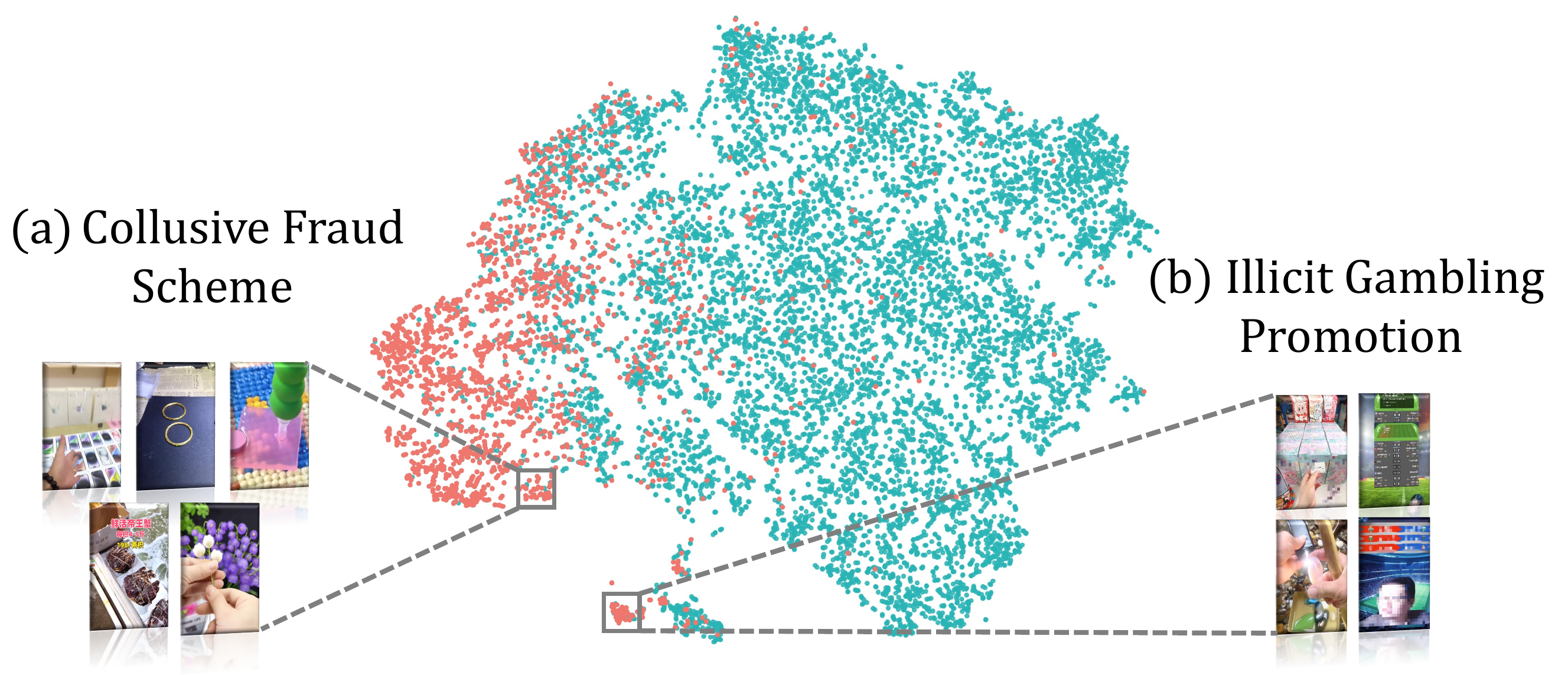}
    \caption{t-SNE visualization of session representations learned by CS-VAR. We highlight two representative clusters: \textbf{collusive fraud schemes} (e.g., fake prize draws, low-cost jewelry/seafood sales, and deceptive part-time jobs) and \textbf{illicit gambling promotion} (e.g., blind-box betting, sports betting, and stone gambling).}
    \label{fig:tsne}
\end{figure}

To address \textbf{RQ4}, we visualize the session representations learned by CS-VAR using t-SNE. We observe the emergence of clusters where sessions differ greatly in surface content yet share recurring risky patterns. For instance, sessions advertising giveaways, merchandise sales, or part-time opportunities align as collusive fraud, while another cluster unifies diverse gambling-related streams under different disguises. These findings reaffirm our motivation: cross-session reasoning equips CS-VAR to identify recurring risky schemes, which reflects the ``déjà vu'' patterns central to live streaming risk assessment.

\subsection{Online Deployment~(RQ5)}

To address \textbf{RQ5}, we have evaluated CS-VAR online to demonstrate its effectiveness.
%, with the detailed deployment workflow illustrated in Appendix~\ref{sec:app:deployment}.
In the deployed environment, we monitored its performance by directly collecting and analyzing real-world traffic processed by the system. As reported in Table~\ref{tab:exp:online}, CS-VAR consistently surpasses both incumbent online models, one based on XGBoost and the other on a Transformer architecture. It achieves substantially higher PR-AUC and F1-score, improves constrained recall by a large margin (0.8305 vs.\ 0.6321 and 0.7830), and reduces false-positive rate to 0.1905. These results confirm that CS-VAR not only outperforms prior deployed solutions but also delivers more reliable and practical detection quality under real production traffic.

\begin{table}[t]
\centering
\caption{Comparison between CS-VAR and incumbent deployed models on real-world production traffic. Metrics are computed on logs spanning 12/18/25–12/19/25 with a 1:10 positive-to-negative sampling ratio.}

\label{tab:exp:online}
\resizebox{0.9\columnwidth}{!}{
  \begin{tabular}{@{}c|cccc@{}} 
  \toprule
  \textbf{} & \textbf{PR-AUC $\uparrow$} & \textbf{F1-score $\uparrow$} & \textbf{R@0.1FPR $\uparrow$} & \textbf{FPR@0.9R $\downarrow$} \\ 
  \midrule
  CS-VAR & \textbf{0.6643} & \textbf{0.6546} & \textbf{0.8305} & \textbf{0.1905} \\
  \midrule
 XGBoost & 0.4543 & 0.4599 & 0.6321 & 0.4481 \\
 Transformer & 0.6096 & 0.5881 & 0.7830 & 0.2173 \\
  \bottomrule
  \end{tabular}
}
\end{table}
\section{Conclusion}
In this paper, we present Cross-Session Evidence-Aware Retrieval-Augmented Detector~(CS-VAR), a framework for live streaming risk assessment that unifies a lightweight, domain-specific model with an LLM to leverage cross-session behavioral evidence. By distilling the LLM’s local-to-global insights into the small model, CS-VAR performs accurate, context-aware risk assessment at the session level while remaining efficient for real-time deployment. Extensive offline evaluations demonstrate state-of-the-art performance, and online validation confirms its practical utility. Beyond accurate detection, CS-VAR generates interpretable signals for moderation, showcasing the value of combining retrieval-augmented LLM reasoning with lightweight models in large-scale industrial applications.
%\begin{acks}
%To Robert, for the bagels and explaining CMYK and color spaces.
%\end{acks}

\appendix

\section{Notation Table}
\label{sec:app:notation}
Here we display a complete list of symbols and corresponding definitions in Table~\ref{tab:notation}.
\begin{table}[h]
\centering

%\footnotesize 
\scriptsize 
\setlength{\tabcolsep}{3pt} 
\caption{Notation table for symbols used in CS-VAR.}
\label{tab:notation}
\begin{tabularx}{\columnwidth}{lX} 
\toprule
\textbf{Symbol} & \textbf{Description} \\
\midrule
$\alpha = (u, t, a, x)$ & An action in a live streaming session: user, timestamp, action type, text embedding \\
$S_s^{[0,T]}$ & Session $s$ as an ordered sequence of actions within $[0,T]$ \\
$N_s$ & Number of actions in session $s$ \\
$\mathcal{U}_s = \{ u_s^{\mathrm{h}} \} \cup U_s^{\mathrm{v},[0,T]}$ & Set of users in session $s$, including host $u_s^{\mathrm{h}}$ and viewers \\
$\mathcal{T}_k = [(k-1)\Delta t, k\Delta t)$ & $k$-th timeslot in temporal discretization, $\Delta t = T/K$ \\
$p_{u,k}$ & Patch: sequence of actions by user $u$ in timeslot $\mathcal{T}_k$ \\
$\mathcal{P}_s$ & Patch grid of session $s$: all $p_{u,k}$ for $u \in \mathcal{U}_s, k=1..K$ \\
$\mathbf{e}_i$ & Embedded action token: concatenation of action-type embedding and projected text embedding \\
$\mathbf{h}_i$ & Contextualized action token after Transformer encoding \\
$\mathbf{p}_{u,k}$ & Patch embedding obtained by LSTM over tokens in patch $p_{u,k}$ \\
$\mathcal{R} = \{ \mathrm{t}, \mathrm{u}, \mathrm{r}, \mathrm{a} \}$ & Set of patch relation types: temporal, user, role, auxiliary \\
$\mathbf{A}^{(\zeta)}$ & Adjacency matrix for relation $\zeta \in \mathcal{R}$ \\
$\mathbf{A}^{\mathrm{G}}$ & Relation-aware adjacency matrix (weighted sum of $\mathbf{A}^{(\zeta)}$) \\
$\mathbf{H}_s$ & Session embedding (from $[\texttt{CLS}]$ token) \\
$\hat{y}_s$ & PatchNet predicted session-level risk score \\
$\hat{y}_k$ & PatchNet predicted patch-level risk score for patch $k$ \\
$\mathcal{P}^{\texttt{key}}_s$ & Selected key patches from session $s$ (top host/viewer patches) \\
$\mathbf{p}_k$ & Embedding of key patch $k$ (query for retrieval) \\
$\mathcal{I}_p$ & Patch index storing embeddings $\mathbf{p}_j$ and session-aware summaries $\tau_j$ \\
$y^{\texttt{LLM}}_k$ & LLM predicted patch-level risk score for key patch $k$ \\
$\texttt{SAL}_k$ & LLM predicted patch-level saliency weight for key patch $k$ \\
$y_s^{\texttt{LLM}}$ & LLM predicted session-level risk score \\
$\tilde{y}_s$ & Weighted aggregation of PatchNet patch-level predictions according to LLM saliency \\
$\mathcal{L}_{\mathrm{s}}$ & Session-level BCE loss \\
$\mathcal{L}_{\mathrm{p}}$ & Patch-level distillation loss \\
$\mathcal{L}_{\mathrm{p2s}}$ & Patch-to-session consistency loss \\
$\mathcal{L}$ & Total loss for cross-granularity distillation \\
\bottomrule
\end{tabularx}

\end{table}
\section{Baselines}
\label{app:baselines}
To rigorously evaluate the performance of our proposed framework, we compare it against a diverse set of established models, which can be broadly partitioned into sequence-based architectures and Instance Aggregation frameworks.

\paragraph{Sequence Models} This group focuses on capturing the temporal dependencies within session action sequences. Specifically, we include the vanilla \textbf{Transformer}~\cite{vaswani2017attention} as a primary benchmark. We further evaluate \textbf{Reformer}~\cite{kitaev2020reformer}, which utilizes locality-sensitive hashing, and \textbf{Informer}~\cite{zhou2021informer}, which leverages ProbSparse attention and distilling techniques.

\paragraph{Instance Aggregation Methods} The second category focuses on session-level inference through instance-to-bag aggregation. Traditional methods like \textbf{mi-NET}~\cite{wang2018revisiting} provide a score-level aggregation baseline. In terms of attention-driven pooling, we incorporate \textbf{AtMIL}~\cite{ilse2018attention} for soft-weighting and \textbf{AdMIL}~\cite{javed2022additive} for its dual emphasis on instance importance and presence. Furthermore, we include contemporary models such as \textbf{MIL-LET}~\cite{early2024inherently}, which employs a decoupled attention-classification strategy, and temporal-aware variants including \textbf{TimeMIL}~\cite{chen2024timemil} (utilizing wavelet tokens) and \textbf{TAIL-MIL}~\cite{jang2025tail} (targeting multivariate time-series via 2D pooling).
\section{Prompt Specification}
\label{sec:app:prompt}
Below, we provide the detailed session-aware summarization prompt employed in Section~\ref{sec:met:index}:
\begin{scriptsize} % 降低一级字体到 footnotesize
\begin{tcolorbox}[
  breakable,
  colback=gray!5, colframe=black!50,
  title=Session-Aware Summarization Prompt,
  boxsep=0pt,          % 消除盒内全局边距
  left=3pt, right=3pt, % 极窄左右边距
  top=2pt, bottom=2pt, % 极窄上下边距
  middle=2pt,          % 标题与内容间的间距
  toptitle=1pt, bottomtitle=1pt, % 标题栏内部间距
  before upper=\renewcommand{\baselinestretch}{0.9}\selectfont % 压缩行间距
]
\setlength{\parskip}{0pt} % 消除段落间距

You are a senior risk control expert with ten years of experience countering underground black-market operations. You understand that malicious actors rarely expose themselves directly. Instead, they use a variety of covert tactics to carry out fraud, gambling, and sexual-content redirection.

Below are multiple user behavior patches from the same live stream (not exhaustive). Each patch represents the sequence of actions by a single user (the host or a viewer) within a 100-second time window. These patches originate from the same live session and may include carefully designed coordinated behavior.

Your task: act like a real risk control expert. From subtle traces, detect anomalies and identify behavior patterns that appear normal on the surface but are suspicious.

\textbf{Guidelines for identifying covert risks:}

1. \textbf{Covert fraud behaviors}: Scripted language masking using words such as ``benefits'', ``discounts'', ``insider information'' instead of explicit scam language; temporal dispersion to avoid concentrated operations; content dilution hiding one or two key guiding messages inside normal content.

2. \textbf{Covert gambling behaviors}: Euphemism system using codewords like ``drive'', ``get on'', ``arrive'' to refer to gambling; platform redirect to third-party platforms before gambling; game disguises such as ``game trials'', ``predictions'', ``mystery box openings'', ``stone gambling'', ``cockfighting''.

3. \textbf{Sexual-content redirection behaviors}: Implicit marketing like ``special services'', ``private benefits'', ``one-on-one''; multi-platform coordination with hints in live session but actions elsewhere; identity packaging as ``model'', ``host assistant'', or ``agent''; time-lag operations building trust then gradually guiding users toward prohibited content.

4. \textbf{Behavior pattern recognition}: Abnormal frequency, timing, or action combinations forming a violation chain.

5. \textbf{Language pattern recognition}: Over-enthusiasm to quickly build trust, creating urgency with ``limited time'' or ``limited slots'', authority packaging by impersonating official staff or professionals.

\textbf{Analysis requirements}: For each patch, perform in-depth analysis (beyond literal text), recognize patterns, analyze correlations between patches, and provide a risk score 0.0–1.0 with detailed explanation.

\textbf{Input format}
\begin{verbatim}
{
  "session_id": "xxx",
  "patches": [
    {"patch_id": 1, "patch_desc": "host spoke 4 times at 00:29: ..."},
    {"patch_id": 2, "patch_desc": "host ..."}
  ]
}
\end{verbatim}

\textbf{Strict output format}
\begin{verbatim}
{
  "session_id": "session_12345",
  "patches": [
    {"patch_id": 1, "risk_score": 0.8, "explanation": "..."},
    {"patch_id": 2, "risk_score": 0.4, "explanation": "..."}
  ],
  "session_summary": "...",
  "overall_risk_score": 0.9,
  "primary_risk_type": "fraud",
  "coordination_indicators": true
}
\end{verbatim}

\textbf{The live streaming session for this query is provided below: < Query Session Content >}

\end{tcolorbox}
\end{scriptsize}

Then, the detailed evidence-integrated reasoning prompt introduced in Section~\ref{sec:met:llm} is shown below:
\begin{scriptsize} 
\begin{tcolorbox}[
  breakable,
  colback=gray!5, colframe=black!50,
  title=Session-Aware Summarization Prompt,
  boxsep=0pt,          % 消除盒内全局边距
  left=3pt, right=3pt, % 极窄左右边距
  top=2pt, bottom=2pt, % 极窄上下边距
  middle=2pt,          % 标题与内容间的间距
  toptitle=1pt, bottomtitle=1pt, % 标题栏内部间距
  before upper=\renewcommand{\baselinestretch}{0.9}\selectfont % 压缩行间距
]
\setlength{\parskip}{0pt} % 消除段落间距

You are a senior live streaming risk control expert with ten years of experience countering underground black-market operations. Malicious actors rarely expose themselves directly; they use covert tactics for fraud, gambling, or sexual-content redirection.

Below are multiple user behavior patches from the same live stream (not exhaustive). Each patch represents the actions of a single user within a 100-second time window, potentially with coordinated behavior.

\textbf{Your task}: Independently analyze the query patch, compare with AI-summarized similar patches, and produce a strict JSON judgment.

\textbf{Analysis Principles}: 
1) Independent analysis of the query patch always comes first; 
2) AI summaries are only secondary references; 
3) Base final decisions on the query patch, but note differences with similar patches; 
4) Assign a saliency score (0.0–1.0) for each patch reflecting weight in overall risk, based on behavioral chain, frequency, coordination, and violation patterns.

\textbf{Key Risk Indicators}:
Fraud: scripted disguises (e.g., ``benefits'', ``discounts''), content dilution, temporal dispersion. 
Gambling: coded language (``drive'', ``get on'', ``arrive''), platform redirection, game disguises (``mystery boxes'', ``predictions'', ``stone gambling''). 
Sexual redirection: implicit marketing (``special services'', ``one-on-one''), multi-platform operation, identity packaging (``model'', ``assistant'', ``agent''). 
Behavioral/language patterns: abnormal frequency, timing, or combinations; over-enthusiasm, urgency, authority signals; coordinated behavior among patches.

\textbf{Input format}
\begin{verbatim}
{
  "session_id": "xxx",
  "patches": [
    {"patch_id": 1, "query_patch": "...", "similar_patch_ai_summary": "..."},
    {"patch_id": 2, "query_patch": "...", "similar_patch_ai_summary": "..."}
  ]
}
\end{verbatim}

\textbf{Output format (strict JSON)}
\begin{verbatim}
{
  "session_id": "xxx",
  "patches": [
    {"patch_id": 1, "risk_score": 0.0, "saliency": 0.0, "explanation": "..."}
  ],
  "session_summary": "...",
  "overall_risk_score": 0.0,
  "primary_risk_type": "fraud | gambling | sexual | normal",
  "coordination_indicators": false
}
\end{verbatim}

\textbf{Reminders}: Independent analysis has priority. Explanations must distinguish independent findings vs. similar patch references. Focus on patterns, abnormal frequencies, and potential coordination. Use behavioral chain logic (actions $\to$ results $\to$ risks). Strictly follow output constraints: `overall\_risk\_score` 0.0–1.0, `primary\_risk\_type` from [normal, fraud, gambling, sexual], `saliency` 0.0–1.0 per patch.

\textbf{The live streaming session for this query is provided below: < Query Session Content >}

\end{tcolorbox}
\end{scriptsize}

\iffalse
\section{Deployment Workflow}
Figure~\ref{fig:deployment} illustrates how CS-VAR is deployed in our production moderation system. 
On the online path, live streaming sessions generate user interaction events that are ingested by a distributed Flink cluster, which aggregates and transforms them into structured logs. These logs are then delivered to a cloud engine that hosts CS-VAR. Within this engine, CS-VAR performs real-time risk prediction, and the outputs are immediately forwarded to the enforcement system for downstream moderation actions. 

In parallel, the offline path leverages the same cloud engine to collect logs (X) together with enforcement outcomes (Y), producing labeled data at scale. This data is used to update CS-VAR, enabling the deployed model to adapt to evolving user behaviors and emerging risk patterns. By separating low-latency prediction from resource-intensive retraining, the architecture ensures both real-time responsiveness and long-term robustness.

\label{sec:app:deployment}
\begin{figure}[htbp]
    \centering
    \includegraphics[width=\linewidth]{figs/deployment.pdf}
\caption{Deployment workflow of CS-VAR in production. 
A distributed Flink cluster streams user interaction logs into the cloud engine, 
where CS-VAR performs real-time risk prediction and forwards results to the enforcement system. 
Offline, the cloud engine aggregates logs (X) with enforcement feedback (Y) to retrain CS-VAR, ensuring continual adaptation to new risks.}
\label{fig:deployment}
\end{figure}

\fi

\begin{acks}
%To Robert, for the bagels and explaining CMYK and color spaces.
The research work is supported by the National Natural Science Foundation of China under Grant Nos. U2436209, 62576333, and 62406307, the Strategic Priority Research Program of the Chinese Academy of Sciences under Grant No. XDB0680201, the Beijing Natural Science Foundation (F251001), and the Innovation Funding of ICT, CAS under Grant No. E461060.
\end{acks}

\bibliographystyle{ACM-Reference-Format}
\bibliography{sample-base}


\end{document}